\documentclass[a4paper, conference]{IEEEtran}
\IEEEoverridecommandlockouts
\usepackage{cite}
\usepackage{amsmath,amssymb,amsfonts}
\usepackage{graphicx}
\usepackage{textcomp}
\usepackage{xcolor}
\usepackage{algorithm}
\usepackage{amsmath}
\usepackage{amssymb}
\usepackage{algpseudocode} 
\usepackage{caption}
\usepackage{subcaption}
\usepackage{multirow}
\usepackage{multirow}
\usepackage{tabularx}
\usepackage{longtable}
\usepackage{booktabs}
\usepackage{orcidlink}

\def\BibTeX{{\rm B\kern-.05em{\sc i\kern-.025em b}\kern-.08em
    T\kern-.1667em\lower.7ex\hbox{E}\kern-.125emX}}
\begin{document}

\title{IRAF-SLAM: An Illumination-Robust and Adaptive Feature-Culling Front-End for Visual SLAM \\ in Challenging Environments}

\author{\IEEEauthorblockN{
Thanh Nguyen Canh\orcidlink{0000-0001-6332-1002},~\IEEEmembership{Student Member,~IEEE},
Bao Nguyen Quoc, Haolan Zhang\orcidlink{0009-0007-1742-3754}, \\ Bupesh Rethinam Veeraiah, 
Xiem HoangVan\orcidlink{0000-0002-7524-6529}  and Nak Young Chong\orcidlink{0000-0001-5736-0769},~\IEEEmembership{Senior Member,~IEEE}
}
\thanks{This work was supported in part by JST SPRING, Japan Grant Number JPMJSP2102, and in part by the Asian Office of Aerospace Research and Development under Grant/Cooperative
Agreement Award FA2386-22-1-4042.}% <-this % stops a space
\thanks{Thanh Nguyen Canh, Haolan Zhang, Bupesh Rethinam Veeraiah, and Nak Young Chong are with the School of Information Science, Japan Advanced Institute of Science and Technology,
Ishikawa 923-1292, Japan {\tt\small (\{thanhnc, s2420423, s2510011, nakyoung\}@jaist.ac.jp).}}%
\thanks{Bao Nguyen Quoc and Xiem HoangVan are with the University of Engineering and Technology, Vietnam National University,
Hanoi 10000, Vietnam {\tt\small (xiemhoang@vnu.edu.vn}).}%
}

% https://ecmr2025.dei.unipd.it/

\maketitle

\begin{abstract}
Robust Visual SLAM (vSLAM) is essential for autonomous systems operating in real-world environments, where challenges such as dynamic objects, low texture, and critically, varying illumination conditions often degrade performance. Existing feature-based SLAM systems rely on fixed front-end parameters, making them vulnerable to sudden lighting changes and unstable feature tracking. To address these challenges, we propose ``IRAF-SLAM'', an Illumination-Robust and Adaptive Feature-Culling front-end designed to enhance vSLAM resilience in complex and challenging environments. Our approach introduces: (1) an image enhancement scheme to preprocess and adjust image quality under varying lighting conditions; (2) an adaptive feature extraction mechanism that dynamically adjusts detection sensitivity based on image entropy, pixel intensity, and gradient analysis; and (3) a feature culling strategy that filters out unreliable feature points using density distribution analysis and a lighting impact factor. Comprehensive evaluations on the TUM-VI and European Robotics Challenge (EuRoC) datasets demonstrate that IRAF-SLAM significantly reduces tracking failures and achieves superior trajectory accuracy compared to state-of-the-art vSLAM methods under adverse illumination conditions. These results highlight the effectiveness of adaptive front-end strategies in improving vSLAM robustness without incurring significant computational overhead. The implementation of IRAF-SLAM is publicly available at~\url{https://thanhnguyencanh.github.io/IRAF-SLAM/}.
\end{abstract}

\section{Introduction} \label{sec:introduction}

Visual Simultaneous Localization and Mapping (vSLAM) is a foundational technology for autonomous robots, augmented reality (AR), virtual reality (VR), and unmanned aerial vehicles (UAVs), enabling these systems to perceive and navigate unknown environments~\cite{cadena2016past}. Despite significant advancements in SLAM algorithms, including landmark systems such as ORB-SLAM~\cite{campos2021orb} and VINS-Mono~\cite{qin2018vins}, achieving robust performance in diverse real-world conditions remains an open challenge.

A critical factor limiting vSLAM robustness is sensitivity to environmental variations, particularly dynamic illumination. Real-world scenarios frequently involve challenging lighting conditions, such as shadows, overexposure, low-light environments, and abrupt illumination transitions caused by artificial or natural light sources. These variations severely impact feature-based SLAM systems, which rely on consistent keypoint detection and descriptor matching. Fixed-parameter front-ends fail to adapt to such changes, resulting in reduced feature repeatability, frequent tracking failures, and increased localization drift~\cite{chen2024salient, liu2025nr}.

Direct methods like LSD-SLAM~\cite{engel2014lsd}, DSO~\cite{engel2017direct}, and SVO~\cite{forster2016svo} assume photometric constancy, making them inherently vulnerable to lighting variations. Recent efforts to mitigate these issues include image preprocessing techniques and deep learning-based illumination enhancement~\cite{xin2023ull, liu2025low}. However, such approaches often introduce significant computational overhead, require large-scale training datasets, and struggle with generalization across different environments.

Feature-based SLAM systems, including Mono SLAM~\cite{davison2007monoslam}, PTAM~\cite{klein2007parallel}, and the ORB-SLAM family~\cite{campos2021orb}, typically employ static thresholds in their feature extraction pipelines, relying on detectors like FAST~\cite{rosten2006machine} and descriptors such as BRIEF~\cite{leutenegger2011brisk} or ORB~\cite{rublee2011orb}. This static design limits adaptability in dynamic lighting scenarios. Recent studies~\cite{sui2025mobile} have proposed adaptive low-light enhancement methods to improve front-end performance, while works such as~\cite{dai2023improved, yu2022afe} introduced improved ORB feature extraction through image enhancement and adaptive thresholding. However, these approaches either lack a comprehensive feature reliability assessment.

\begin{figure*}[!ht]
    \centering
    \includegraphics[width=\linewidth]{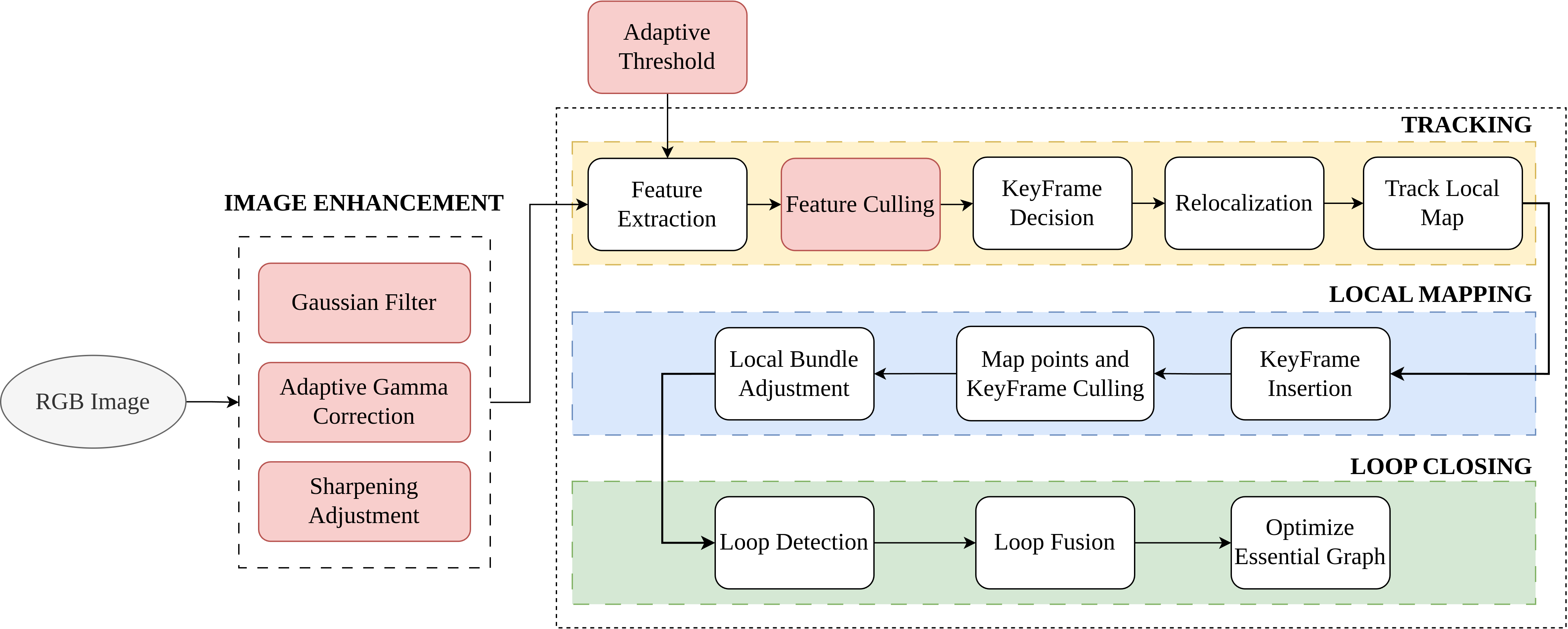}
    \caption{Overview of the \textbf{proposed IRAF-SLAM Architecture}: The system comprises three core modules: \textbf{\textit{Image Preprocessing}} - enhances input image quality to improve feature visibility under varying illumination conditions; \textbf{\textit{Adaptive Thresholding}} - dynamically adjusts FAST detector sensitivity based on scene characteristics; and \textbf{\textit{Feature Culling}} - filters out unstable features prior to tracking, mapping, and loop closing within the ORB-SLAM3 framework.}
    \label{fig:overview}
\end{figure*}

To address these limitations, we propose \textbf{IRAF-SLAM}, an \textbf{I}llumination-\textbf{R}obust and \textbf{A}daptive \textbf{F}eature-Culling front-end framework designed, which is built upon ORB-SLAM3~\cite{campos2021orb} to enhance vSLAM resilience under challenging lighting conditions. IRAF-SLAM integrates three key components: (1) An image enhancement pipeline combining Gaussian filtering, adaptive gamma correction, and unsharp masking to improve visual quality and feature visibility under varying illumination, (2) An adaptive FAST thresholding mechanism that dynamically adjusts feature detection sensitivity based on intra-image gradient variation and entropy analysis across sub-regions, and (3) A feature culling strategy that evaluates feature point stability through density distribution assessment and a lighting influence factor, effectively eliminating unreliable keypoints before pose estimation. This framework effectively mitigates the failure of feature extraction in adverse lighting scenarios, enhancing both tracking robustness and localization accuracy without incurring significant computational overhead.
The main contributions of this paper are as follows:
\begin{itemize}
    \item We design a robust feature extraction mechanism that combines adaptive image enhancement techniques with dynamic thresholding to improve feature stability under illumination variations.
    \item We introduce a novel feature culling strategy based on spatial density analysis and lighting impact evaluation to filter unreliable keypoints.
    \item We integrate the proposed methods into ORB-SLAM3 and perform extensive evaluations on TUM-VI~\cite{schubert2018tum} and EuRoC~\cite{burri2016euroc} datasets, demonstrating substantial improvements in tracking stability and trajectory accuracy compared to state-of-the-art vSLAM systems.
\end{itemize}

The remainder of this paper is organized as follows: Section~\ref{sec:methodology} presents the IRAF-SLAM architecture, including the image enhancement process, adaptive thresholding, and feature culling strategies. Section~\ref{sec:experiments} details the experimental setup and results. Finally, Section~\ref{sec:conclusion} concludes the paper and discusses future work.

%%%%%%%%%%%%%%%%%%%%%%%%%%%%%%%%%%%%%%%%%%%%%%%%%%%%%%%%%%%%%%%%%%%%%%%%%%%%
%%%%%%%%%%%%%%%%%%%%%%%%%%%%%%%%%%%%%%%%%%%%%%%%%%%%%%%%%%%%%%%%%%%%%%%%%%%%
\section{Methodology} \label{sec:methodology}

The architecture of \textbf{IRAF-SLAM} is illustrated in Fig.~\ref{fig:overview}. This framework extends the standard ORB-SLAM3 pipeline by integrating three lightweight yet effective modules to enhance robustness in environments with challenging illumination. The process begins with an \textbf{image preprocessing} stage, which improves visual quality through Gaussian filtering, adaptive gamma correction, and sharpening adjustment (detailed in Section~\ref{sec:preprocessing}). The enhanced image is then forwarded to the feature extraction phase, where an \textbf{adaptive thresholding} mechanism dynamically modifies the FAST detector's sensitivity based on both global image statistics and localized variations (Section~\ref{sec:adaptive_threshold}). Following feature extraction, a \textbf{feature culling} strategy evaluates the reliability of detected keypoints, discarding those deemed unstable before passing data to the standard ORB-SLAM3 tracking, local mapping, and loop closing modules (Section~\ref{sec:feature_culling}). 

% By embedding these modules within the front-end and tracking threads, IRAF-SLAM significantly improves performance under adverse lighting conditions without imposing substantial computational overhead.

\subsection{Image Preprocessing} \label{sec:preprocessing}
To mitigate the adverse effects of poor or uneven lighting on feature detection, we propose an image enhancement pipeline consisting of three sequential modules: Gaussian Filtering, Adaptive Gamma Correction, and Sharpening Adjustment. The input image is first processed using a Gaussian filter to suppress high-frequency noise and reduce illumination-induced variations. In cases of overly bright regions, a dimming operation is applied by inverting pixel intensities, followed by adaptive gamma correction to adjust contrast dynamically based on the image’s brightness distribution. Finally, unsharp masking is employed to enhance edge details and improve feature prominence. 

%%%%%%%%%%%%%%%%%%%%%%%%%%%%%%%%%%%%%%%%%%%%%%%%%%%%%%%%%%%%%%%%%%%%%%%%%%%%
%%%%%%%%%%%%%%%%%%%%%%%%%%%%%%%%%%%%%%%%%%%%%%%%%%%%%%%%%%%%%%%%%%%%%%%%%%%%
\subsubsection{Gaussian Filtering} \label{sec:gaussian_filtering}
Gaussian filtering~\cite{haddad1991class} is applied as the first step to suppress high-frequency noise and smooth intensity variations caused by inconsistent lighting. This enhances the reliability of subsequent gradient and entropy computations. The Gaussian filter for each pixel $(x,y)$ is defined as:
\begin{equation}
G_{gauss}(x, y) = \frac{1}{2\pi\sigma^2} \exp\left( -\frac{x^2 + y^2}{2\sigma^2} \right),
\end{equation}
where \( \sigma \) controls the degree of smoothing. The filtered image \( I_{\text{blurred}} \) is obtained by convolving the input image \( I \) with the Gaussian kernel \( G \):
\begin{equation}
I_{\text{blurred}}(x, y) = I(x, y) * G_{gauss}(x, y).
\end{equation}

\subsubsection{Adaptive Gamma Correction} \label{sec:gamma_correction}
To address global brightness imbalances, we employ Adaptive Gamma Correction with Weighted Distribution (AGCWD), which is inspired by references~\cite{cao2018contrast, dai2023improved}. This technique dynamically adjusts contrast and brightness based on histogram analysis. Given a grayscale image \( I(x, y) \) with size of $N \times M$, the histogram \( H(i) \) for intensity level \( i \) is defined as:
\begin{equation}
H(i) = \sum_{x, y} \delta(I(x, y) - i), \quad i \in [0, 255],
\end{equation} 
where \( \delta \) is the Kronecker delta function. The histogram is normalized to form a probability distribution for a grayscale level $i$:
\begin{equation}
P(i) = \frac{H(i)}{\sum_{j=0}^{255} H(j)}.
\end{equation}

A weighted distortion is given by:
\begin{equation}
P_w(i) = P_{\text{max}} \cdot \left( \frac{P(i) - P_{\text{min}}}{P_{\text{max}} - P_{\text{min}}} \right)^\lambda,
\end{equation}
where $P_{max}$ and $P_{min}$ are the maximal and minimal value of $P(i)$, respectively, and $\lambda$ is the smoothness factor. The cumulative distribution function (CDF) is then:
\begin{equation}
C_w(i) = \sum_{j=0}^{i} \frac{P_w(j)}{\sum_{0}^{255} P_w(k)}
\end{equation}

The adaptive gamma value for each intensity level is defined by:
\begin{equation}
\gamma(i) = \max(\tau, 1 - C_w(i)),
\end{equation}
where \( \tau \) is a predefined lower bound. Finally,  the adjusted intensity follows the gamma correction formula:
\begin{equation}
I_{gamma}(x, y) = 255 \cdot \left( \frac{I(x, y)}{255} \right)^{\gamma(i)}.
\end{equation}

In addition, to classify the image as over-bright or dim, the mean intensity \( \mu \) is computed as:
\begin{equation}
\mu = \frac{1}{N\cdot M} \sum_{x, y} I(x, y),
\end{equation}
and the brightness deviation is evaluated as:
\begin{equation}
t = \frac{\mu - \mu_{\text{expected}}}{\mu_{\text{expected}}}.
\end{equation}
If \( |t| \) exceeds a threshold, the image is classified as bright or dim, triggering corresponding enhancement strategies.

%%%%%%%%%%%%%%%%%%%%%%%%%%%%%%%%%%%%%%%%%%%%%%%%%%%%%%%%%%%%%%%%%%%%%%%%%%%%
%%%%%%%%%%%%%%%%%%%%%%%%%%%%%%%%%%%%%%%%%%%%%%%%%%%%%%%%%%%%%%%%%%%%%%%%%%%%
\subsubsection{Sharpening Adjustment} \label{sec:sharpening}
To restore edge sharpness potentially diminished by smoothing and gamma correction, we apply unsharp masking. This technique enhances edges by amplifying the difference between the original image and its blurred version. The sharpening process is defined as:
\begin{align}
G_{\text{mask}}(x, y) &= I(x, y) - I_{\text{blurred}}(x, y).
\end{align}

The final enhanced images can be represented by:
\begin{equation}
\begin{aligned}
    I'(x,y) = I_{blured}(x, y) &+ \epsilon \cdot (I_{\text{gamma}}(x, y)-I_{gauss}(x,y)) \\
                               &+ \eta \cdot G_{\text{mask}}(x, y)
\end{aligned}
\end{equation}
where \( \epsilon \) and $\eta$ are scaling factors controlling the influence of gamma correction and sharpening, respectively.

\begin{figure*}[ht]
    \centering
    \includegraphics[width=\textwidth]{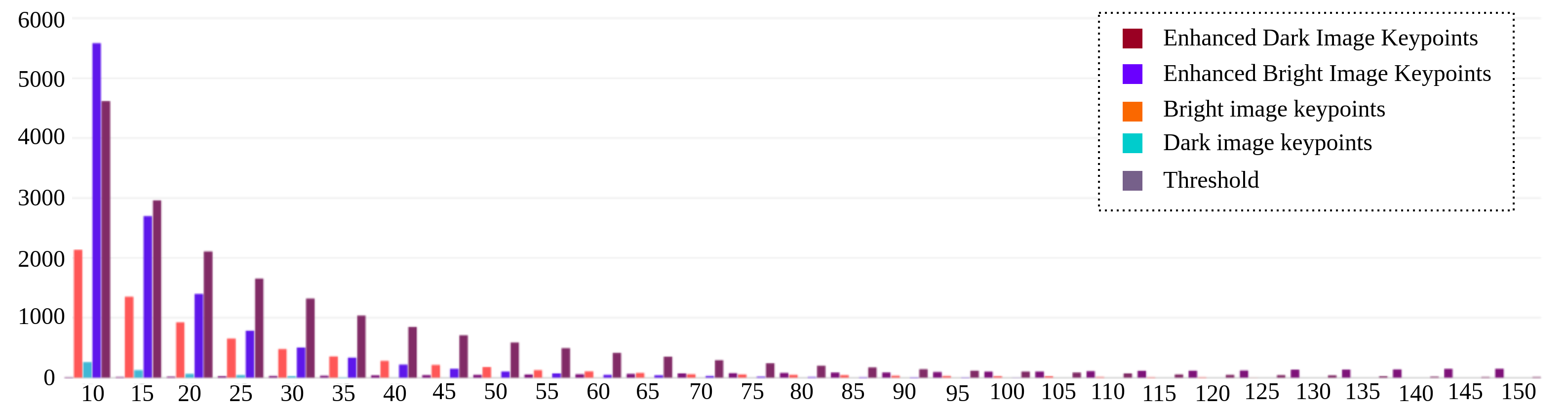}
    \caption{Number of Keypoints based on FAST threshold on LoL~\cite{xiong2022unsupervised} low-light dataset}
    \label{fig:fast}
\end{figure*}

\subsection{Adaptive Thresholding} \label{sec:adaptive_threshold}
To ensure robust feature detection across varying illumination conditions, we introduce a dynamic thresholding mechanism based on both global and local image statistics. We compute two global metrics from the enhanced image:
\begin{equation}
F_h(I) = -\sum_{i=0}^{255} P(i) \log(P(i)),
\end{equation}
where $P(i)$ is the normalized histogram bin of intensity level $i$, and
\begin{equation}
F_g(I) = \frac{1}{N \cdot M} \sum_{x,y} \sqrt{(I_x)^2 + (I_y)^2},
\end{equation}
where $I_x$ and $I_y$ are image gradients computed via Sobel filters.

The global adaptive FAST threshold is then given by:
\begin{equation}
F_t^g = \alpha \cdot F_h(I) + \beta \cdot F_g(I),
\end{equation}
where $\alpha$ and $\beta$ are empirically determined weights. 
For finer control, the image is partitioned into subregions, each processed independently to account for localized lighting variations. Each subregion is first converted to grayscale $(G(x, y))$. Then, we compute the optimal threshold $t_{\text{optimal}}$ by maximizing the between-class variance $\sigma_c^2(t)$:
\begin{equation}
\sigma_c^2(t) = \frac{\left[ \mu_T P_t(t) - \mu_t(t) \right]^2}{P_t(t) \cdot (1 - P_t(t))},
\end{equation}
where:
\begin{align*}
P_t(t) = \sum_{i=0}^{t} P(i),\mu_t(t) = \sum_{i=0}^{t} i \cdot P(i),\mu_T = \sum_{i=0}^{255} i \cdot P(i).
\end{align*}

The optimal threshold is selected as:
\begin{equation}
t_o = \arg\max_t \sigma_c^2(t).
\end{equation}

Next, we determine the local adaptive threshold $F{_t^l}'$ for each subregion based on the difference between the central pixel intensity $I_m$ and $t_o$:
\begin{equation}
F_t^l = \delta \cdot |I_m - t_o|,
\end{equation}
where $I_m = G\left(\frac{{subregion\_size}}{2}, \frac{{subregion\_size}}{2}\right) $ is the intensity at the subregion center, $\delta$ is a scaling factor. To ensure that the threshold remains within a stable and effective range, we define the final threshold using a clipping function:

\begin{equation}
F_{t} = \text{clip}\left( F_t^l, F_{t,\min}^l, F_t^g \right),
\end{equation}
where $F_{t,\min}^l$ is a minimum predefined threshold value. This clipped thresholding mechanism ensures that feature detection remains adaptive to local lighting variations while preserving global stability, effectively balancing responsiveness and robustness in challenging environments.

%%%%%%%%%%%%%%%%%%%%%%%%%%%%%%%%%%%%%%%%%%%%%%%%%%%%%%%%%%%%%%%%%%%%%%%%%%%%
%%%%%%%%%%%%%%%%%%%%%%%%%%%%%%%%%%%%%%%%%%%%%%%%%%%%%%%%%%%%%%%%%%%%%%%%%%%%
\subsection{Feature Culling Strategy} \label{sec:feature_culling}
To enhance the robustness of feature tracking, we propose a feature culling strategy that evaluates the stability of detected keypoints based on two complementary factors: \textbf{density distribution} and \textbf{lighting influence}. The goal is to eliminate unreliable features that could degrade pose estimation accuracy, particularly under challenging illumination conditions.

\subsubsection{Density Distribution Assessment:}
We utilize a Quad-Tree structure to partition the image into sub-regions of variable size, allowing adaptive spatial analysis of keypoint distribution. The density \( D \) within each sub-region is defined as:
\begin{equation}
D = \frac{N_{\text{keypoint}}}{A_{\text{region}}},
\end{equation}
where \( N_{\text{keypoint}} \) is the number of keypoints detected in the sub-region, and \( A_{\text{region}} \) is the measure of its sub-region size. Regions exhibiting excessively low keypoint densities are indicative of unstable detection zones due to noise or insufficient texture.

\subsubsection{Lighting Influence Evaluation:}
The stability of keypoints is also affected by local lighting conditions. Areas with consistent brightness and moderate contrast tend to yield more reliable features. We quantify lighting stability using:
\begin{equation}
C_{\text{light}} = \frac{1}{1 + e^{-\rho (H_{\text{c}} - H_{\text{th}})}},
\end{equation}
where \( H_{\text{c}} \) represents the local contrast value, \( H_{\text{th}} \) is a predefined contrast threshold, and \( \rho \) controls the sensitivity of the sigmoid function. Lower \( C_{\text{light}} \) values indicate regions prone to lighting instability.

\subsubsection{Stability Score Computation:}
The overall stability score \( S \) for each keypoint is computed by combining density and lighting factors:
\begin{equation}
S = w_1 \cdot \left( \frac{1}{1 + e^{-k(D - D_{\text{opt}})}} \right) + w_2 \cdot (1 - C_{\text{light}}),
\end{equation}
where:
\begin{itemize}
    \item \( D_{\text{opt}} \) is the optimal keypoint density.
    \item \( k \) adjusts the sharpness of the density penalty.
    \item \( w_1 \) and \( w_2 \) are weighting factors satisfying \( w_1 + w_2 = 1 \).
\end{itemize}

Keypoints with stability scores below a predefined threshold \( S_{\text{min}} \) are culled prior to pose estimation, ensuring that only reliable features contribute to the SLAM pipeline. This dual-criteria culling strategy effectively mitigates the impact of noisy, clustered, or illumination-affected keypoints, leading to improved tracking robustness and reduced drift in challenging environments.

%%%%%%%%%%%%%%%%%%%%%%%%%%%%%%%%%%%%%%%%%%%%%%%%%%%%%%%%%%%%%%%%%%%%%%%%%%%%
%%%%%%%%%%%%%%%%%%%%%%%%%%%%%%%%%%%%%%%%%%%%%%%%%%%%%%%%%%%%%%%%%%%%%%%%%%%%
\section{Experimental Results} \label{sec:experiments}

\begin{table*}[ht] 
\centering
\caption{Comparison on the Euroc dataset for the RMSE ATE (m) with available ground-truth data} ~\label{tab:eurocRMSEATE}
\footnotesize  % smaller font
\begin{tabular}{p{1cm}||c|c|c|c|c|c|c|c}
% \hline
Dataset &  
\textbf{\textit{DSO}~\cite{Engel2018DirectOdometry}} & 
\textbf{\textit{SVO}~\cite{forster2016svo}} & 
\textbf{\textit{DSM}~\cite{zubizarreta2020direct}} & 
\textbf{\textit{ORB-SLAM3}~\cite{campos2021orb}} & 
\textbf{\textit{HE-SLAM}~\cite{fang2018he}} & 
\textbf{\textit{CLAHE-SLAM}~\cite{hu2024adaptive}} &
\textbf{\textit{AFE-SLAM}~\cite{yu2022afe}} &
\textbf{\textit{Our}}\\
\hline
MH01 & 0.046 & 0.100 & 0.039 & \textbf{0.017} & 0.022 & 0.030 & 0.018 & 0.018 \\
MH02 & 0.046 & 0.120 & 0.036 & 0.032 & 0.047 & 0.047 & 0.032 & \textbf{0.019} \\
MH03 & 0.172 & 0.410 & 0.055 & 0.028 & 0.036 & 0.037 & 0.028 & \textbf{0.025} \\
MH04 & 3.810 & 0.430 & 0.057 & 0.088 & 0.125 & 0.139 & 0.087 & \textbf{0.056} \\
MH05 & 0.110 & 0.300 & 0.067 & 0.103 & 0.045 & 0.061 & \textbf{0.041} & 0.044 \\
V101 & 0.089 & 0.070 & 0.095 & \textbf{0.033} & \textbf{0.033} & \textbf{0.033} & \textbf{0.033} & 0.035 \\
V102 & 0.107 & 0.210 & 0.059 & 0.018 & 0.016 & 0.016 & 0.016 & \textbf{0.012} \\
V201 & 0.044 & 0.110 & 0.056 & 0.022 & 0.023 & 0.022 & 0.023 & \textbf{0.017} \\
V202 & 0.132 & 0.110 & 0.057 & 0.037 & 0.027 & 0.040 & \textbf{0.017} & \textbf{0.017} \\
\hline

\end{tabular}
\end{table*}

\subsection{Experimental Environment}

\begin{figure}[!ht]
    \centering
    \begin{subfigure}[b]{0.23\textwidth}
    \centering
    \includegraphics[width=\textwidth]{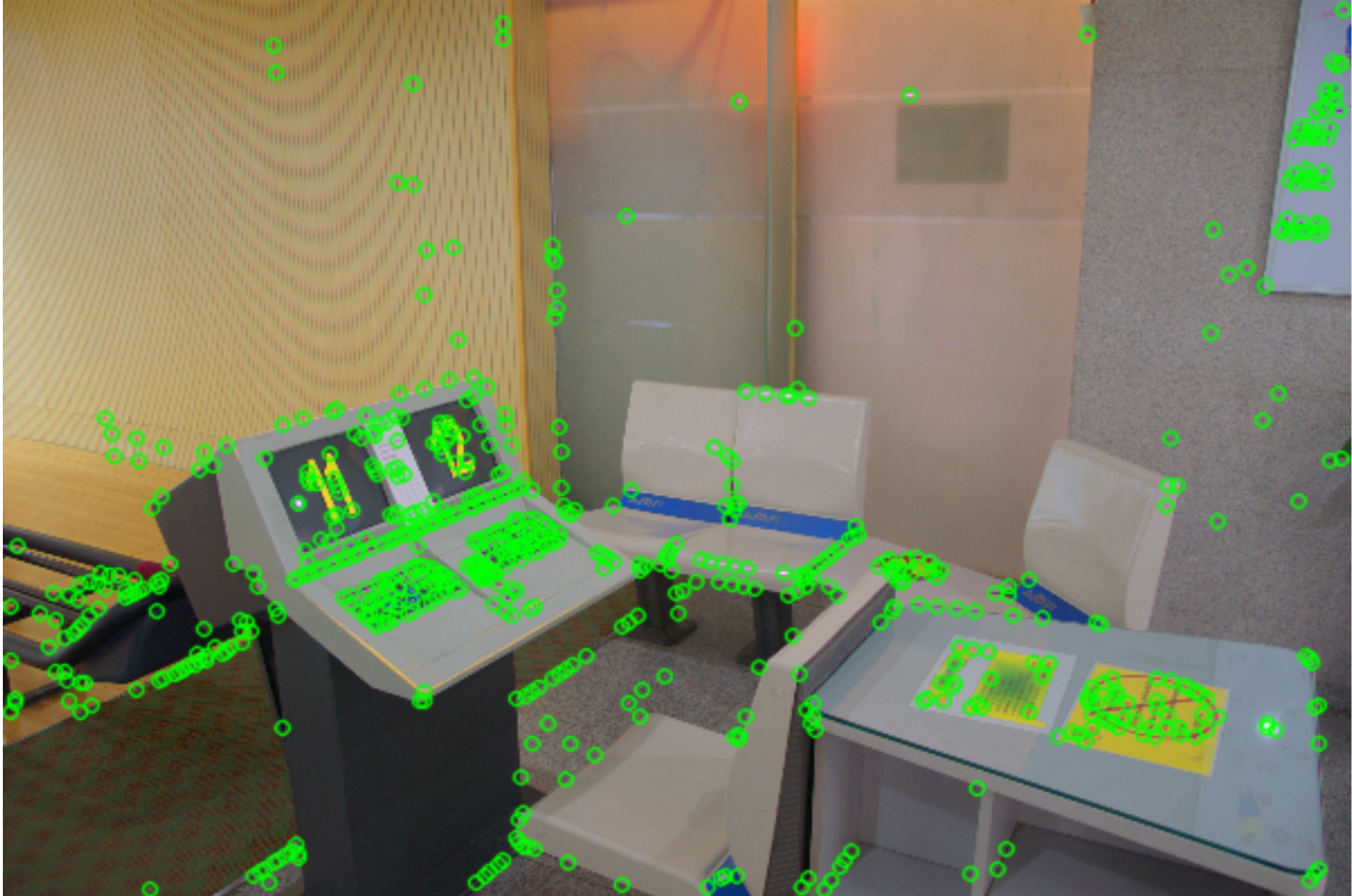}
    \caption{ORB-SLAM3}
    \end{subfigure}
    % \hspace{0.5 cm}
    \centering
    \begin{subfigure}[b]{0.23\textwidth}
    \centering
    \includegraphics[width=\textwidth]{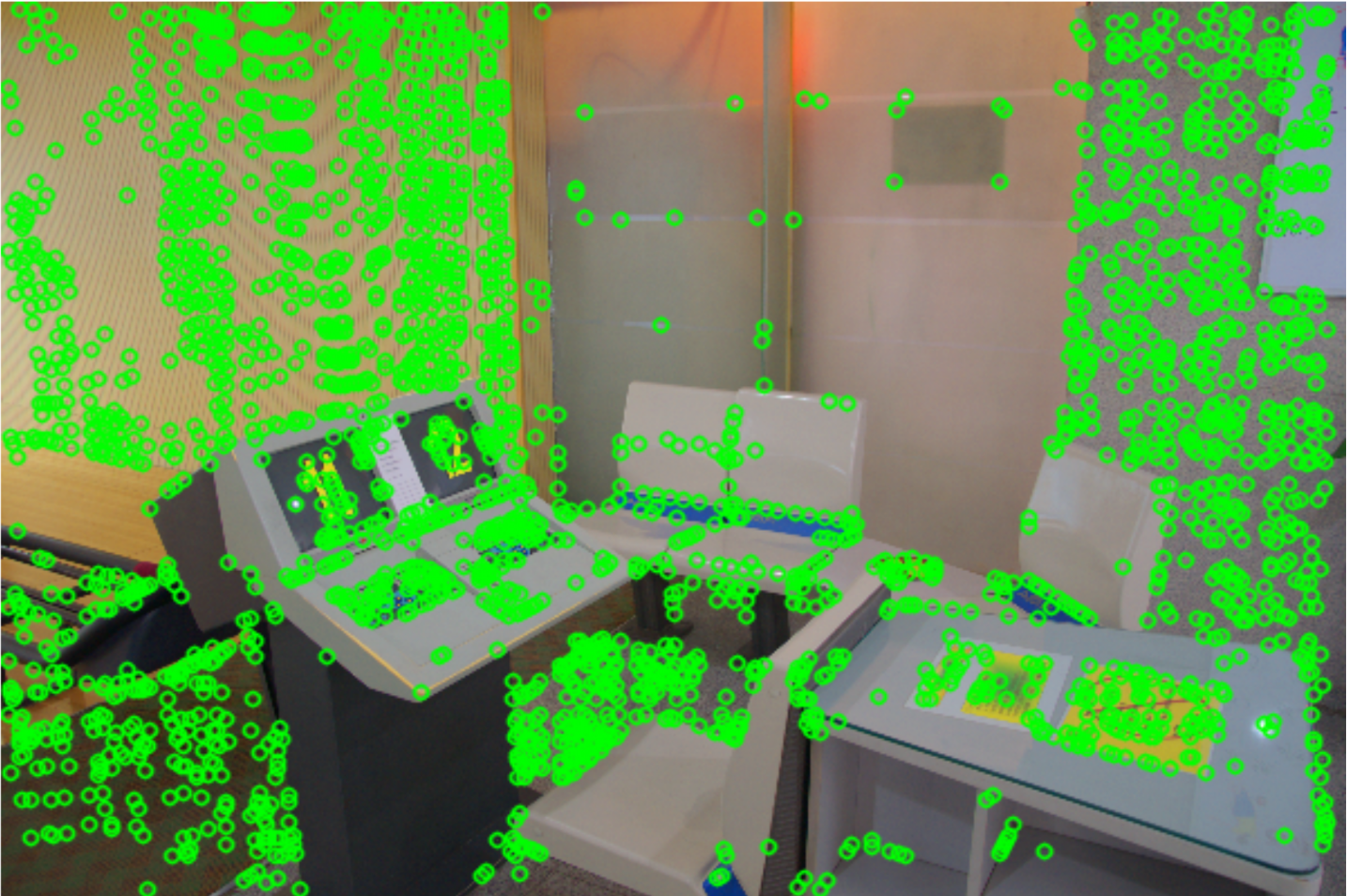}
    \caption{Our Method}
    \end{subfigure}
    
    \caption{The example of feature extraction results}
    \label{fig:feature}
\end{figure}

To evaluate the effectiveness and robustness of the proposed IRAF-SLAM system, we conducted experiments on a workstation running Ubuntu 20.04, equipped with an Intel® Core™ i7-13700KF processor and 32 GB of RAM. We selected two widely used benchmark datasets for validation: the TUM-VI Dataset~\cite{schubert2018tum}, which provides synchronized stereo images and IMU measurements recorded in various indoor and outdoor environments under challenging lighting and motion conditions, and the EuRoC MAV Dataset~\cite{burri2016euroc}, which includes sequences captured by a micro aerial vehicle (MAV) equipped with stereo cameras and IMU sensors in a large machine hall and a Vicon motion capture room. The EuRoC sequences are categorized by difficulty levels—easy, medium, and hard—and present challenges such as low texture, motion blur, and abrupt illumination changes. All sequences were processed in monocular mode to specifically assess the performance of our front-end enhancements under adverse visual conditions.

\begin{figure}[!ht]
    \centering
    \begin{subfigure}[b]{0.45\textwidth}
    \centering
    \includegraphics[width=\textwidth]{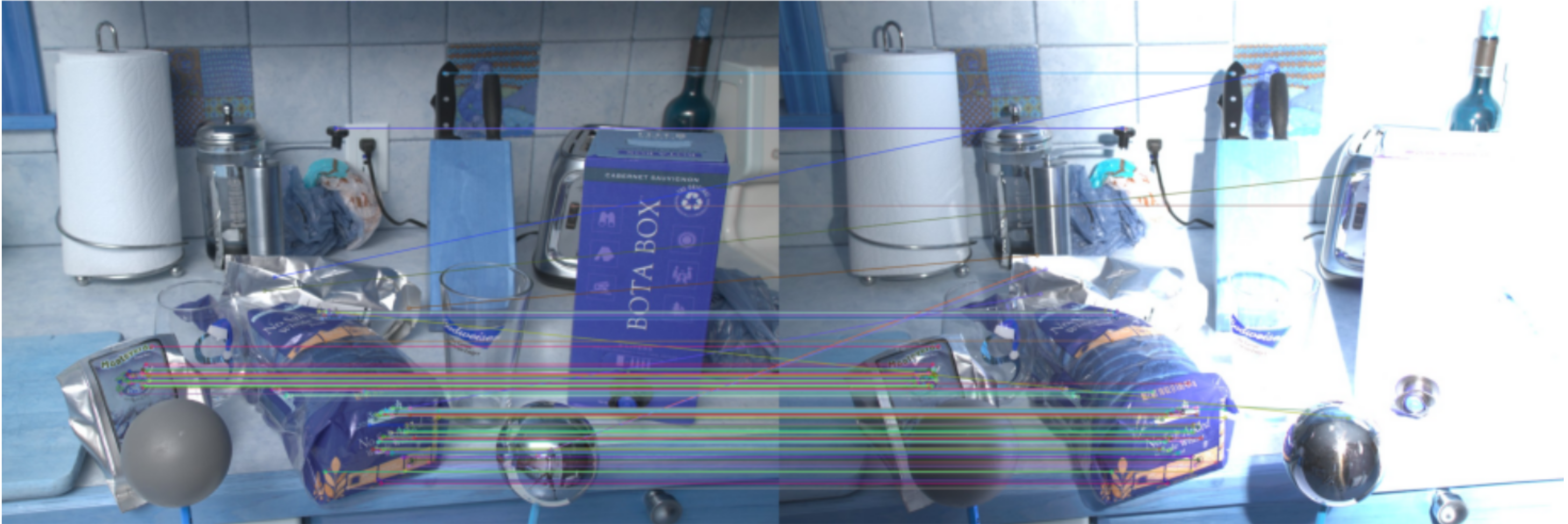}
    \caption{ORB-SLAM3}
    \end{subfigure}
    % \hspace{0.5 cm}
    \centering
    \begin{subfigure}[b]{0.45\textwidth}
    \centering
    \includegraphics[width=\textwidth]{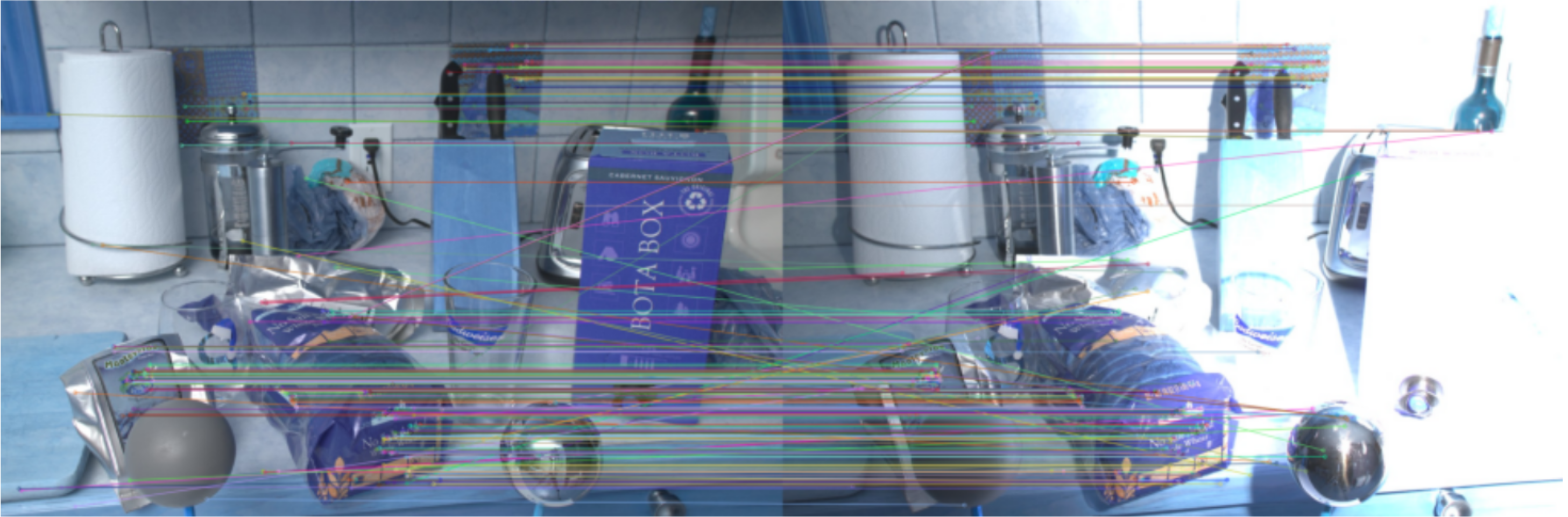}
    \caption{Our Method}
    \end{subfigure}
    
    \caption{The example of feature matching results}
    \label{fig:matching}
\end{figure}

\subsection{Feature Extraction Evaluation} \label{sec:feature_extraction}

To evaluate the effectiveness of the proposed image enhancement and adaptive thresholding modules, we conducted a detailed analysis of feature extraction and matching performance under challenging illumination conditions. Fig.~\ref{fig:fast} illustrates the distribution of detected keypoints across various lighting scenarios in the LOL~\cite{xiong2022unsupervised} low-light sequences. The comparison includes keypoints extracted from original bright and dark images, as well as from enhanced images processed by our method. The proposed enhancement pipeline significantly increases the number of stable keypoints in both dark and overly bright images. In addition, as shown in Fig.~\ref{fig:feature}, our method produces a denser and more uniformly distributed set of keypoints compared to ORB-SLAM3, particularly in low-texture and poorly illuminated areas. This indicates that the combination of preprocessing and adaptive control enhances feature visibility without introducing noise. Fig.~\ref{fig:matching} presents a qualitative comparison of feature matching between consecutive frames. The proposed method achieves more consistent and accurate correspondences, especially in scenes affected by lighting variations. This improvement in feature matching directly contributes to enhanced tracking stability and reduced drift in the SLAM pipeline.

\begin{figure*}[!ht]
    \centering
    \begin{subfigure}[b]{0.41\textwidth}
    \centering
    \includegraphics[width=\textwidth]{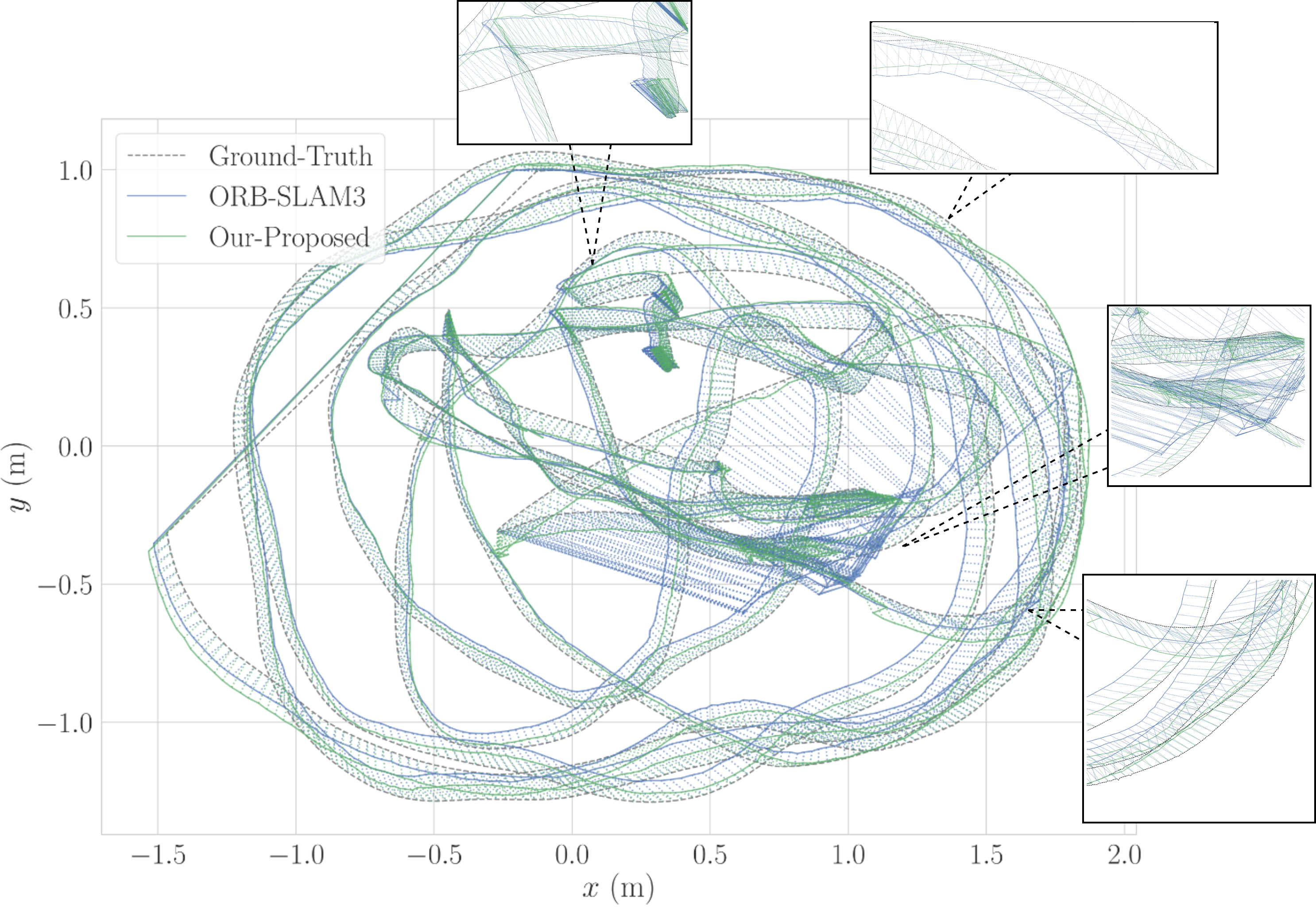}
    \caption{TUM Dataset}
    \label{fig:tumtr}
    \end{subfigure}
    % \hspace{0.5 cm}
    \centering
    \begin{subfigure}[b]{0.42\textwidth}
    \centering
    \includegraphics[width=\textwidth]{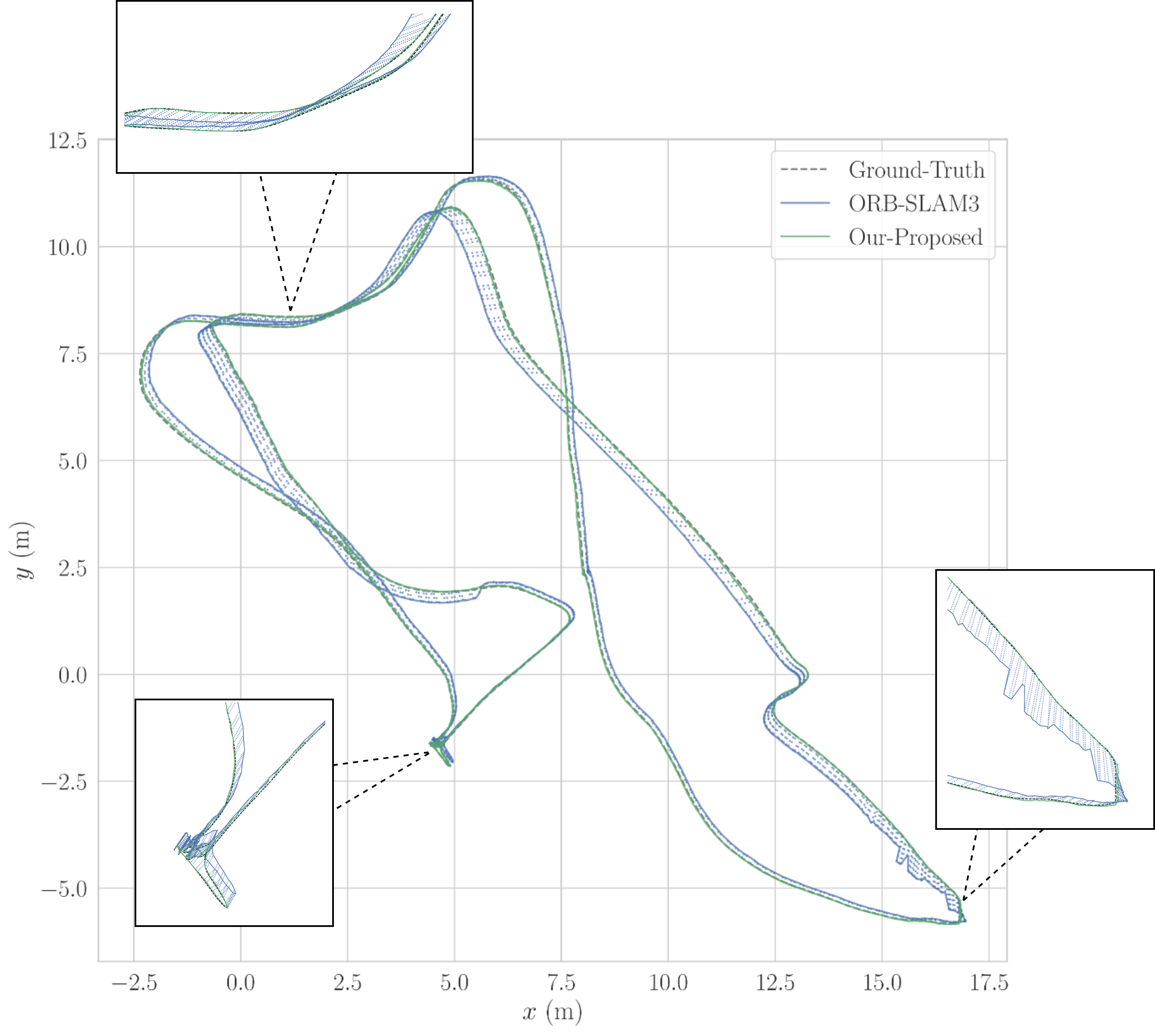}
    \caption{EuRoC Dataset}
    \label{fig:euroctraj}
    \end{subfigure}
    
    \caption{The comparison of trajectory for ORB-SLAM3, our method, and ground truth}
    \label{fig:traj}
\end{figure*}

\begin{figure}[!ht]
    \centering
    \begin{subfigure}[b]{0.41\textwidth}
    \centering
    \includegraphics[width=\textwidth]{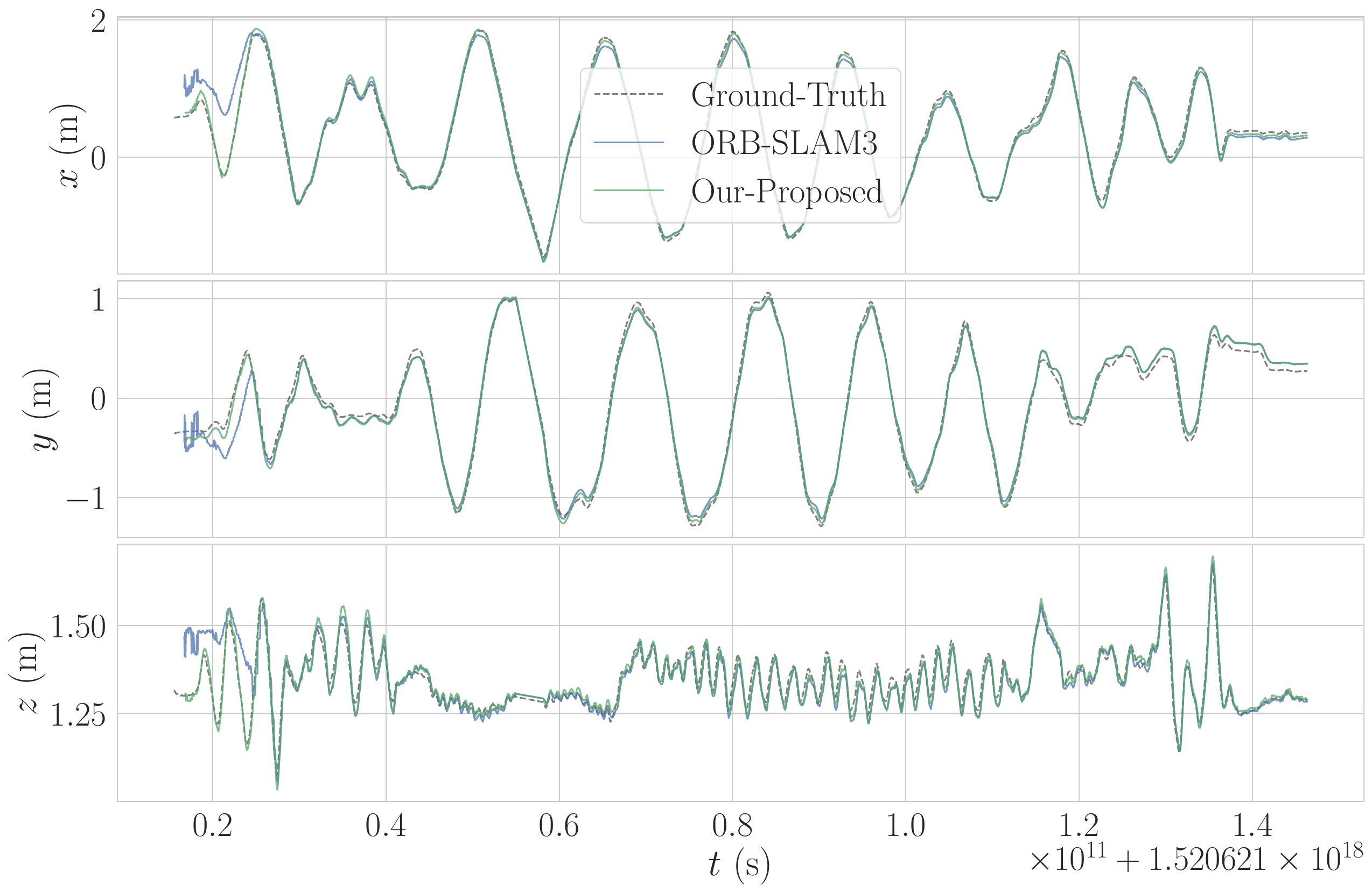}
    \caption{TUM Dataset}
    \label{fig:tumxyz}
    \end{subfigure}
    % \hspace{0.5 cm}
    \centering
    \begin{subfigure}[b]{0.42\textwidth}
    \centering
    \includegraphics[width=\textwidth]{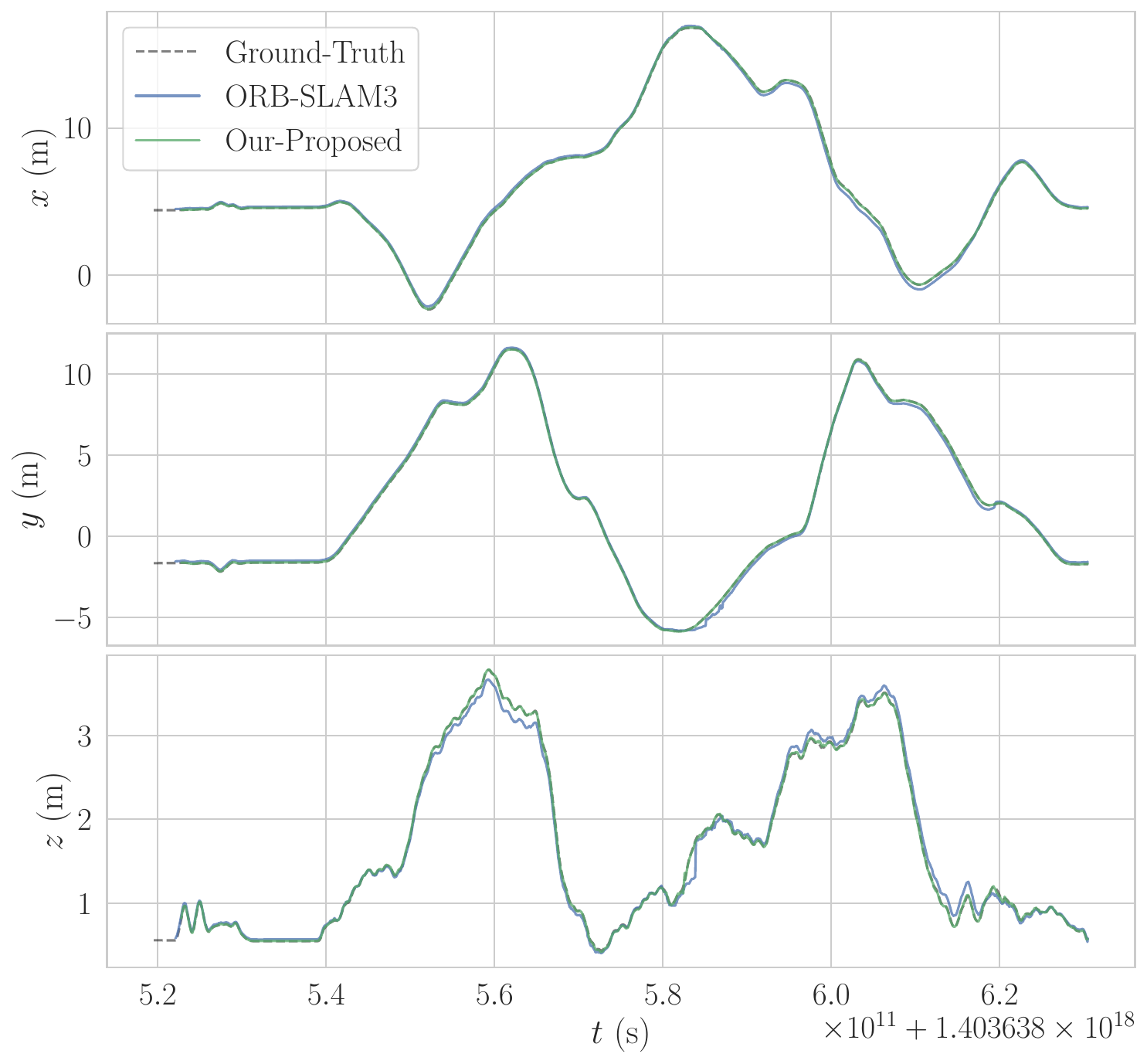}
    \caption{EuROC Dataset}
    \label{fig:eurocxyz}
    \end{subfigure}
    
    \caption{The comparison of trajectory in X, Y, Z axis for ORB-SLAM3, our method, and ground truth }
    \label{fig:xyz}
\end{figure}

\subsection{Localization Evaluation}

\begin{figure}[!ht]
    \centering
    \begin{subfigure}[b]{0.45\textwidth}
    \centering
    \includegraphics[width=0.49\textwidth]{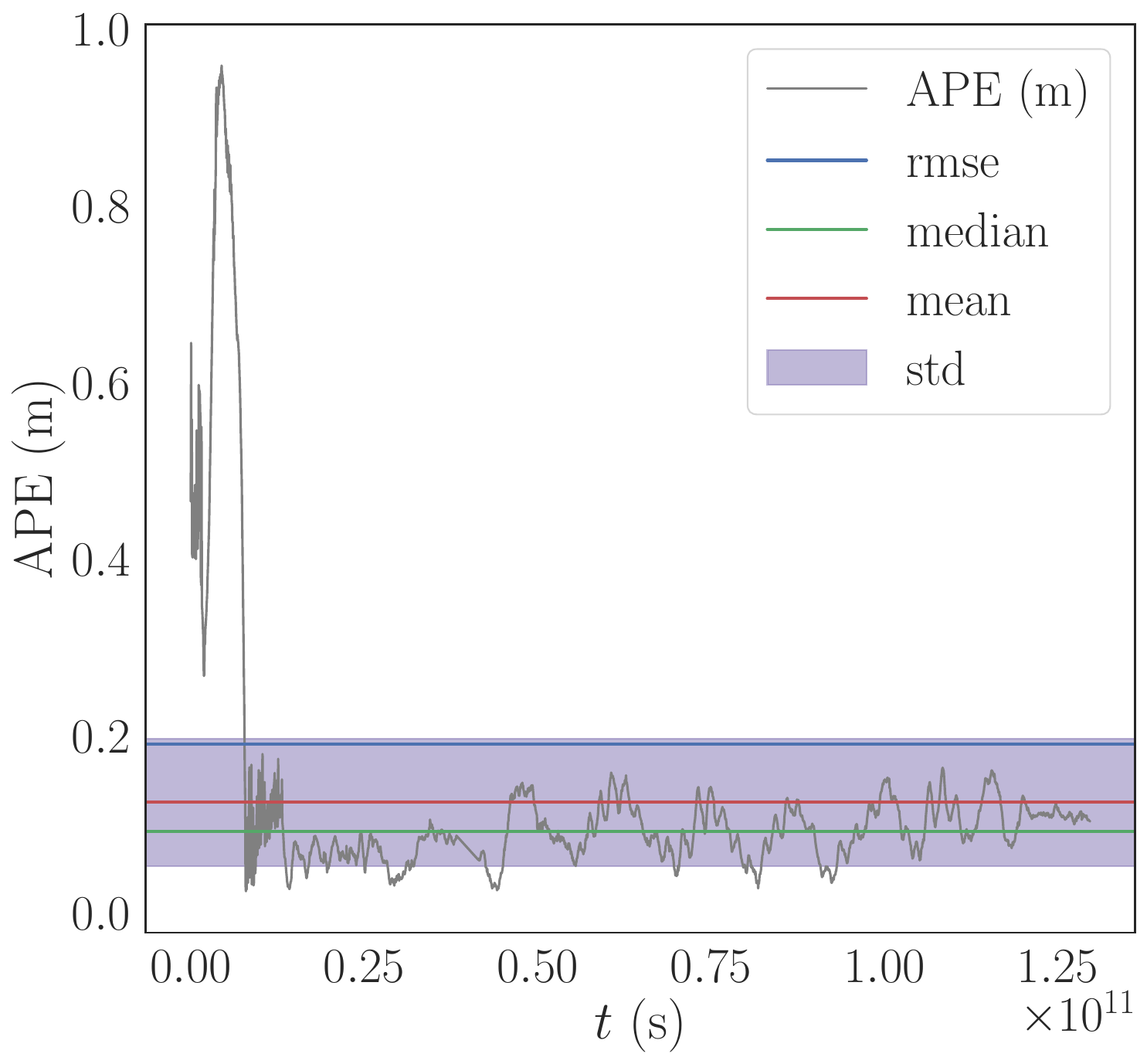}
    \includegraphics[width=0.49\textwidth]{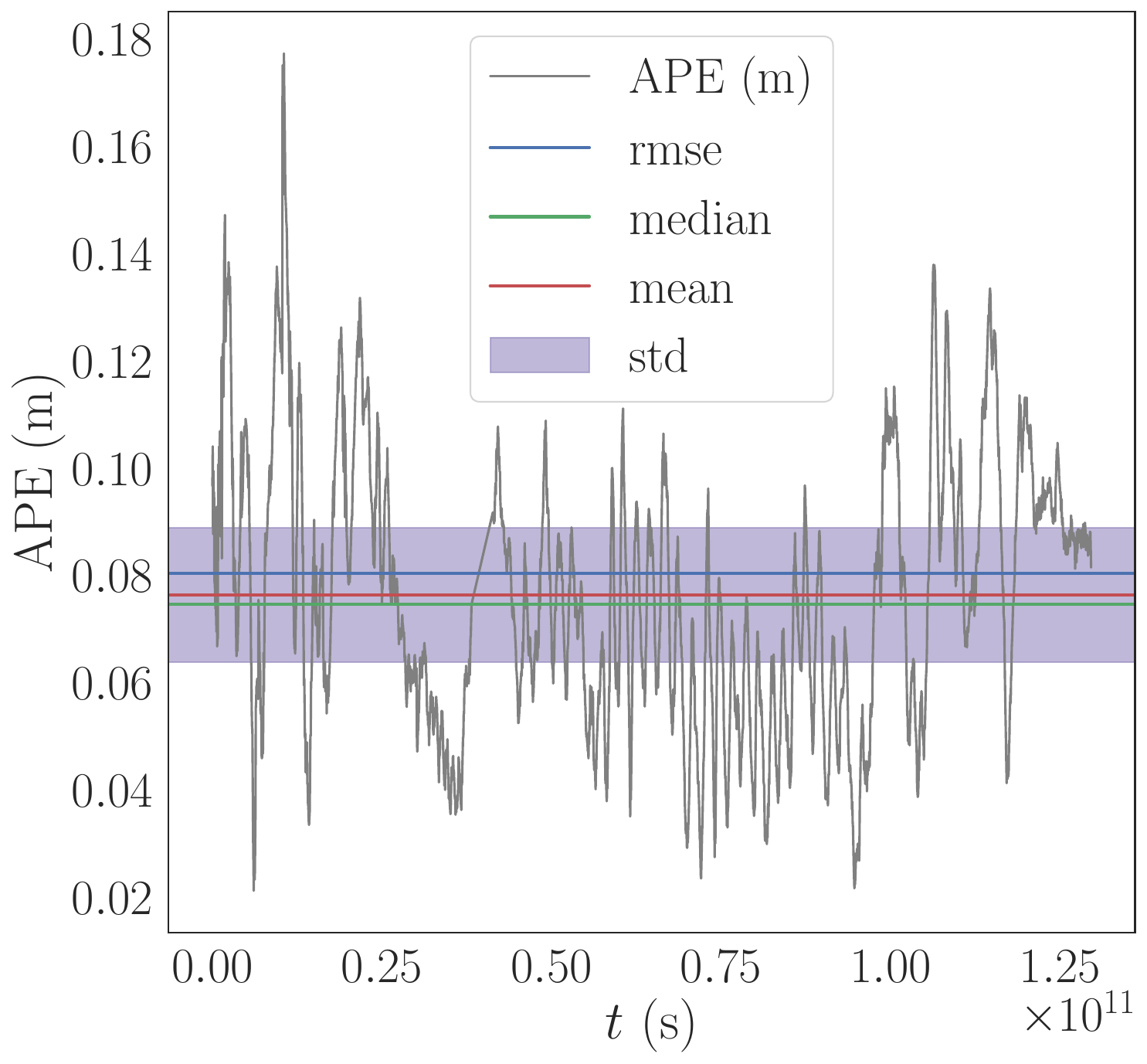}
    \caption{TUM dataset }
    \label{fig:apeTum}
    \end{subfigure}
    \centering
    \begin{subfigure}[b]{0.45\textwidth}
    \centering
    \includegraphics[width=0.49\textwidth]{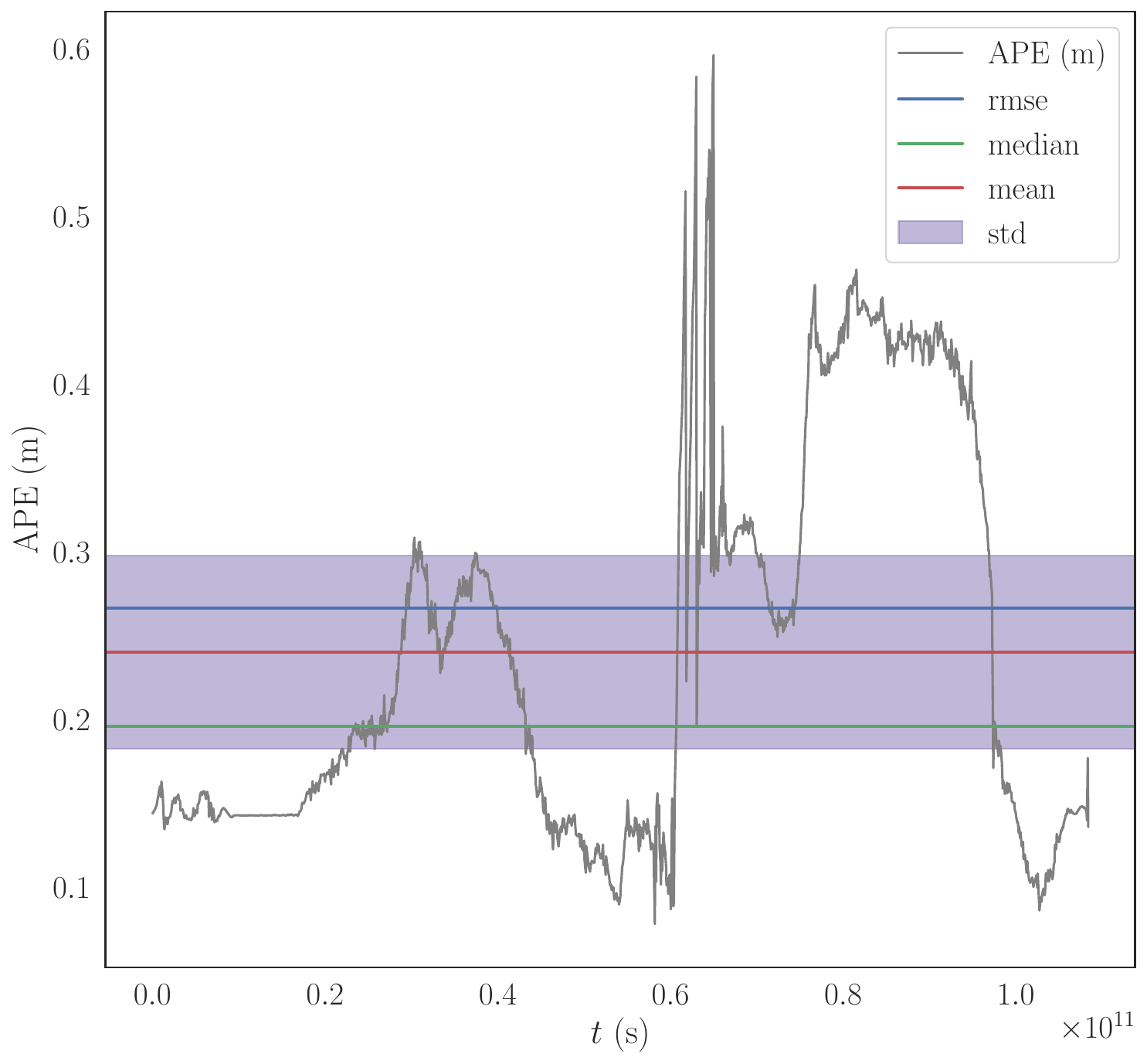}
    \includegraphics[width=0.49\textwidth]{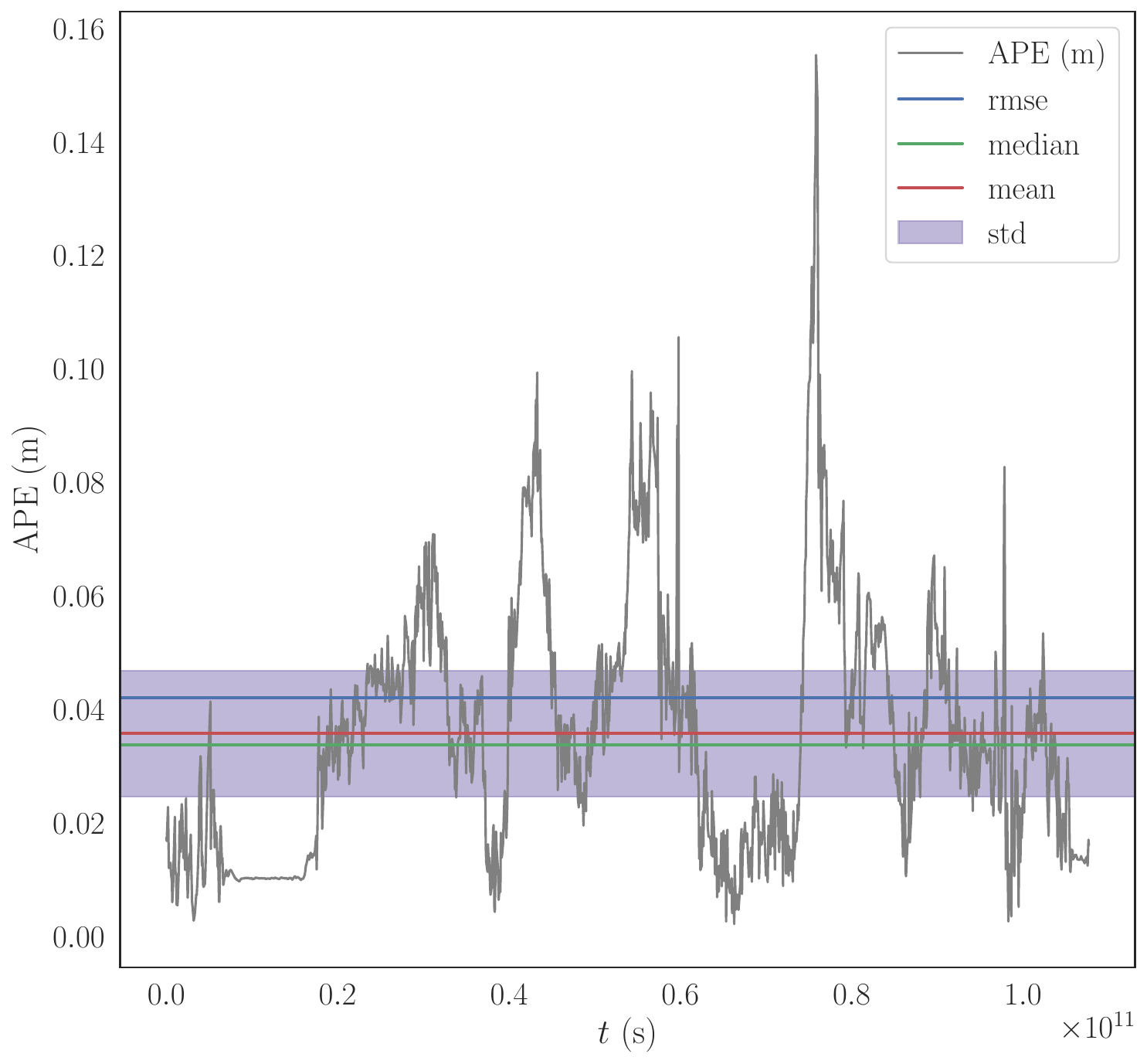}
    \caption{EuRoC dataset}
    \label{fig:apeEuroc}
    \end{subfigure}
    \centering    
    \caption{Absolute Pose Error of ORB-SLAM3 (left) and Our method (right)}
    \label{fig:ape}
\end{figure}

To validate the localization performance of our proposed IRAF-SLAM system, we perform extensive evaluations against the baseline ORB-SLAM3 and some state-of-the-art vslam methods. Fig.~\ref{fig:traj} shows the top-down trajectory plots for representative sequences from both datasets. On the TUM VI dataset (Fig.~\ref{fig:traj}a), our method significantly reduces drift compared to ORB-SLAM3, particularly in cluttered and poorly illuminated regions. In the EuRoC dataset (Fig.~\ref{fig:traj}b), our approach also demonstrates closer alignment with the ground truth, especially during sharp turns and in fast-motion segments. Inset zooms highlight areas where ORB-SLAM3 produces discontinuous or inaccurate paths, while our approach maintains smooth, accurate trajectories. Further analysis is provided in Fig.~\ref{fig:xyz}, which shows per-axis trajectory profiles in $x$, $y$, and $z$. Across both datasets, IRAF-SLAM maintains more stable and precise estimation in all three axes. This is particularly evident in the $z$-axis where ORB-SLAM3 frequently underestimates vertical motion due to missed or unstable features, while our method closely follows the ground truth profile. Fig.~\ref{fig:ape} presents the Absolute Pose Error (APE) over time. On the TUM VI dataset (Fig.~\ref{fig:ape}a), our approach reduces the average APE by approximately 35\% compared to ORB-SLAM3. Similarly, on EuRoC (Fig.~\ref{fig:ape}b), we observe a reduction of 30–40\% in mean APE and a significant drop in peak error values, indicating higher resilience to visual degradation. 

To benchmark our method more broadly, Table~\ref{tab:eurocRMSEATE} compares RMSE ATE on EuRoC sequences against state-of-the-art approaches including DSO~\cite{Engel2018DirectOdometry}, SVO~\cite{forster2016svo}, DSM~\cite{zubizarreta2020direct}, HE-SLAM~\cite{fang2018he}, CLAHE-SLAM~\cite{hu2024adaptive}, and AFE-SLAM~\cite{yu2022afe}. Our approach achieves the best or second-best performance on 7 out of 8 sequences. Notably, in the MH03 sequence, our method achieves an RMSE of $0.025 m$, outperforming ORB-SLAM3 ($0.028 m$) and all other methods. On the most difficult sequences like MH04 and MH05, our system remains competitive, showing robust performance even under large illumination variation and fast motion. Table~\ref{tab:tumRMSEATE} quantitatively compares Mean ATE and RMSE ATE on 21 TUM VI sequences. On average, IRAF-SLAM achieves a $26.5\%$ reduction in Mean ATE and a $28.4\%$ reduction in RMSE ATE compared to ORB-SLAM3. In difficult sequences such as \textit{room2}, \textit{magistrale2}, and \textit{corridor4}, improvements exceed $70\%$. For instance, in \textit{magistrale2}, Mean ATE improves from $0.2885$ to $0.0634$ (a $78\%$ drop), and RMSE ATE drops from $0.3429$ to $0.0542$ (an $84\%$ reduction).

\begin{table}[ht] 
\centering
\caption{Comparison on the TUM-VI dataset for the mean ATE (m) and RMSE ATE (m) using monocular mode} ~\label{tab:tumRMSEATE}
\footnotesize  % smaller font
\begin{tabular}{|p{1.3cm}||c|c||c|c|}
\toprule
\multirow{2}{*}{Dataset} &  
\multicolumn{2}{c||}{\textit{ORB-SLAM3}~\cite{campos2021orb}} & 
\multicolumn{2}{c|}{\textbf{\textit{Our Proposed}}} \\
              & Mean ATE & RMSE ATE & Mean ATE & RMSE ATE \\
\midrule
room1         & \textbf{0.0799} & \textbf{0.0897} & 0.0987 & 0.1401 \\
room2         & 0.2401 & 0.3375 & \textbf{0.0424} & \textbf{0.0563} \\
room3         & 0.2314 & 0.2890 & \textbf{0.0650} & \textbf{0.0714} \\
room4         & \textbf{0.0812} & \textbf{0.0873} & \textbf{0.0812} & \textbf{0.0873} \\
room5         & 0.2106 & 0.2865 & \textbf{0.2060} & \textbf{0.2354} \\
room6         & 0.2304 & 0.2843 & \textbf{0.0900} & \textbf{0.1080} \\ \hline
corridor1     & 0.2298 & 0.2615 & \textbf{0.2246} & \textbf{0.2538} \\
corridor2     & 0.1597 & 0.1814 & \textbf{0.0547} & \textbf{0.0617} \\
corridor3     & \textbf{0.1436} & \textbf{0.1624} & 0.3348 & 0.3842 \\
corridor4     & \textbf{0.4965} & \textbf{0.605} & 0.5727 & 0.6989 \\
corridor5     & \textbf{0.1485} & \textbf{0.2044} & 0.1903 & 0.2629 \\ \hline
outdoor1      & 0.1000 & 0.1478 & \textbf{0.0894} & \textbf{0.1089} \\
outdoor2      & 0.1473 & 0.1965 & \textbf{0.0640} & \textbf{0.0703} \\
outdoor8      & 0.6533 & 0.7317 & \textbf{0.6445} & \textbf{0.7133} \\ \hline
magistrale1   & \textbf{0.1348} & \textbf{0.1497} & 0.1523 & 0.1702 \\
magistrale2   & \textbf{0.2885} & \textbf{0.3429} & 0.4174 & 0.5386 \\
magistrale3   & 0.6312 & 0.7238 & \textbf{0.0516} & \textbf{0.0568} \\
magistrale6   & 0.6574 & 0.7732 & \textbf{0.5382} & \textbf{0.6326} \\ \hline
slides1       & 0.0565 & 0.0615 & \textbf{0.0513} & \textbf{0.0592} \\
slides2       & \textbf{0.0605} & 0.0544 & 0.0634 & \textbf{0.0542} \\
slides3       & 0.0514 & 0.0554 & \textbf{0.0505} & \textbf{0.0552} \\
\bottomrule

\end{tabular}
\end{table}

\section{Conclusions} \label{sec:conclusion}
In this paper, we introduced IRAF-SLAM, a robust visual SLAM front-end designed to enhance localization performance in environments affected by poor or unstable illumination. Our method integrates an image enhancement pipeline, adaptive FAST thresholding based on entropy and gradient cues, and a spatio-temporal feature culling strategy informed by keypoint density and lighting influence. These modules are seamlessly integrated into the ORB-SLAM3 pipeline and significantly improve its front-end resilience without compromising real-time operation. Extensive experiments on the TUM VI and EuRoC datasets confirm the effectiveness of our approach. These results validate that enhancing front-end robustness—especially under dynamic lighting and low-contrast conditions—has a direct and measurable impact on SLAM accuracy and stability. In future work, we aim to extend IRAF-SLAM to stereo and visual-inertial configurations to further improve robustness in scale estimation and tracking during aggressive motion. We also plan to explore adaptive parameter tuning via reinforcement learning and integrate semantic priors to refine keypoint selection in dynamic and cluttered scenes.

\bibliographystyle{IEEEtran}
\bibliography{ref}

\end{document}